\documentclass{article}
\usepackage{spconf,amsmath,graphicx}
\usepackage{makecell}
\usepackage{booktabs}
\usepackage{multicol}
\usepackage{multirow}
\usepackage{graphicx}
\usepackage{amssymb}
\usepackage{tabularx}
\usepackage{array}
\usepackage{url} 
\usepackage{threeparttable}
\usepackage[normalem]{ulem}
\newcolumntype{N}{>{\centering\arraybackslash}m{.53in}}
\newcolumntype{G}{>{\centering\arraybackslash}m{2in}}
\newcolumntype{Y}{>{\centering\arraybackslash}X}

\title{Improving Large-scale Deep Biasing with Phoneme Features and Text-only Data in Streaming Transducer}
\name{Jin Qiu$^*$, Lu Huang$^*$\thanks{*\ Equal contribution.}, Boyu Li, Jun Zhang, Lu Lu, Zejun Ma}
\address{ByteDance\\
    \texttt{\{qiujin,huanglu.thu19\}@bytedance.com}}

\begin{document}
%
\maketitle
\begin{abstract}
Deep biasing for the Transducer can improve the recognition performance of rare words or contextual entities, which is essential in practical applications, especially for streaming Automatic Speech Recognition (ASR). However, deep biasing with large-scale rare words remains challenging, as the performance drops significantly when more distractors exist and there are words with similar grapheme sequences in the bias list. In this paper, we combine the phoneme and textual information of rare words in Transducers to distinguish words with similar pronunciation or spelling. Moreover, the introduction of training with text-only data containing more rare words benefits large-scale deep biasing. The experiments on the Librispeech corpus demonstrate that the proposed method achieves state-of-the-art performance on rare word error rate for different scales and levels of bias lists.

\end{abstract}


\section{Introduction}
\label{sec:intro}

Recently, E2E models have been widely explored in the ASR community and have achieved significant improvements \cite{sainath2020streaming, li2022recent}. Compared to hybrid models, E2E ASR directly maps speech features into word sequences by optimizing a single neural network with E2E criteria. The most popular methods, such as CTC \cite{graves2006connectionist, li2018advancing}, Transducer \cite{sainath2020streaming, li2020developing, graves2012sequence}, and Attention-based Encoder-Decoder (AED) \cite{chorowski2015attention, karita2019comparative, li2020comparison}, have become mainstream.

Since the vocabulary comprises sub-word units, it is difficult for E2E models to recognize rare words, as they are frequently decomposed into infrequent sub-word sequences \cite{fu2023robust}. However, in practical applications, the accurate recognition of rare words is crucial for providing a better user experience, such as in the case of songs, contacts, installed applications. Moreover, rare words and text corpus containing such words are often available in advance. Therefore, finding ways to leverage this information and benefit E2E ASR models has become increasingly important.

One of the most common methods to improve the recognition performance of rare words is language model (LM) fusion. It can be achieved by constructing an FST based on rare words or contextual words \cite{pundak2018deep, zhao2019shallow} and incorporating it during beam search. Besides, an external task-specific LM trained on extra text-only data can be used during inference \cite{allauzen2021hybrid, huang2022sentence}. Also, the external LM can be incorporated in the E2E model during training \cite{peyser2020improving, weiran22_interspeech}, like MWER training \cite{prabhavalkar2018minimum}.

An alternative solution is to bias the E2E model with an all-neural framework. CLAS \cite{pundak2018deep} was proposed to incorporate contextual information dynamically into the E2E model. In \cite{le2021deep}, a deep personalized LM was introduced to influence the model’s predictions earlier, and the performance was further improved by combining shallow fusion (SF), deep biasing, and LM contextualization \cite{le2021contextualized}. CATT \cite{chang2021context, hou2022bring} was proposed by jointly training a context-biasing network with the original Transducer. In \cite{sathyendra2022contextual}, a contextual adapter is added to the pre-trained ASR model, which is conditioned with the outputs of the encoder's different layers \cite{dingliwal2023personalization}. Besides, some works use an auxiliary loss or decoder to predict the rare words directly \cite{huang2023contextualized, han2021cif}. Additionally, the phoneme information of rare words was also considered for deep biasing \cite{bruguier2019phoebe}, and different embedding extractors were explored in \cite{chen2019joint}. 

In addition, there are some works that explore the usage of extra text corpus through text injection, rather than relying on LM. This is achieved by joint training of speech and text data, like JOIST \cite{sainath2023joist,sainath2023improving}, MAESTRO \cite{chen2022maestro}, JEIT \cite{meng2023jeit}. Furthermore, text-only domain adaptation has gained popularity recently, as seen in approaches like TOG \cite{thomas2022integrating} and USTR \cite{huang2023text}.

\begin{figure*}[th]
    \centering
    \includegraphics[width=0.98\textwidth]{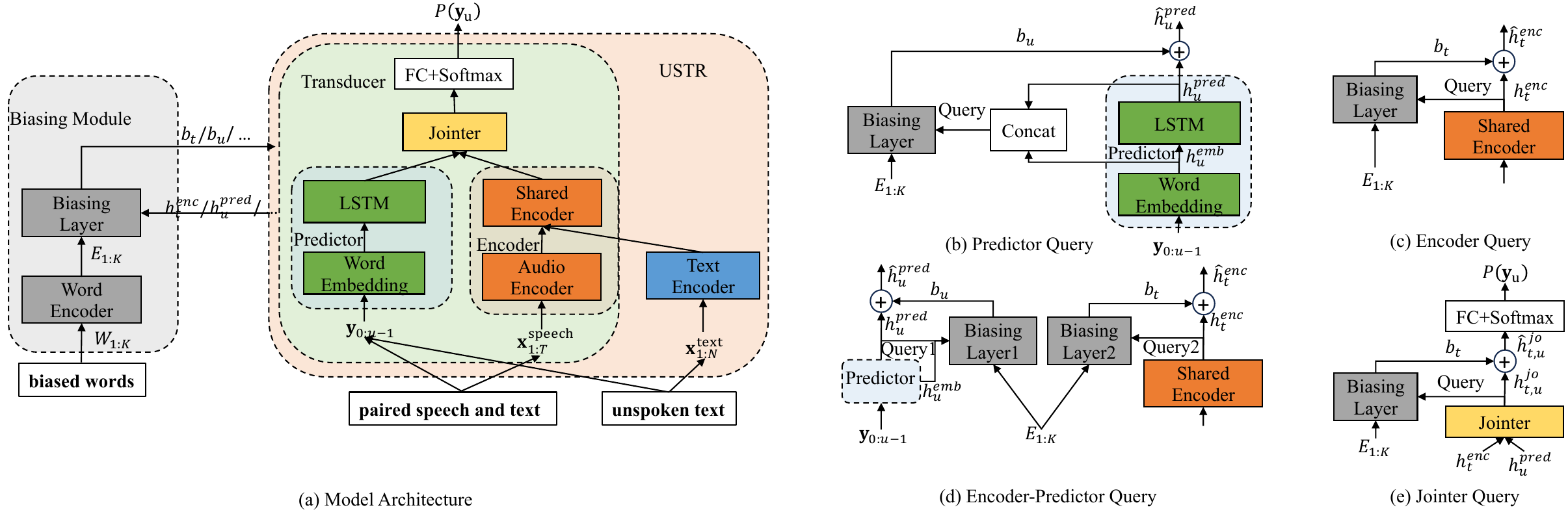}
    \caption{The model structures of proposed method. (a) is the overall architecture. (b) is the case where the query of biasing layer is from \texttt{Predictor}. (c) is \texttt{Encoder Query}. (d) is the \texttt{Enc-Pre Query}. (e) is \texttt{Jointer Query}.}
    \label{fig:model}
\end{figure*}

In this study, we first explores deep biasing modules using different hidden states as condition (query in attention). These modules are trained with a pre-trained Transducer, and the proposed learning rate policy can achieve better accuracy on biased words, and maintain the performance on unbiased words. 
Specifically, encoder-predictor query is chosen for its better performance and lower computational cost. To enhance the performance of deep biasing with large-scale bias lists, the phoneme information of rare words is also combined with the textual information. To our best knowledge, this is the first time that phoneme information is adopted for biasing Transducer. Additionally, we introduced the previous USTR approach to further improve the training of biasing module, and two types of text-only corpora are explored to demonstrate the feasibility of our methodology. Compared to previous methods, our experiments on Librispeech corpus showed that the proposed framework achieves competitive or even superior performance regarding the recognition accuracy of rare words.


\section{Related Work}
\label{sec:related}

\subsection{Deep biasing}
\label{subsec:attention bias}
Attention-based deep biasing methods are most relevant to our work, such as
CLAS \cite{pundak2018deep} and C-RNNT \cite{jain2020contextual}, where an attention mechanism is used to direct the model's focus towards specific contextual entities.  
Meanwhile, in C-RNNT \cite{chang2021context,sathyendra2022contextual}, deep biasing is employed on both the encoder and predictor in Conformer Transducers (CT).
Also, CLAS is further improved with phoneme representations \cite{bruguier2019phoebe,chen2019joint}, and leveraging phoneme information achieves better discrimination for similar grapheme sequences \cite{fu2023robust,chen2019joint}. 

However, these works may increase the decoding complexity without biasing module due to the combiner module. Furthermore, phoneme information hasn't been explored for biasing Transducers before, and we show that phoneme information brings a gain for large scale deep biasing.

Additionally, it is found that optimizing the adapter by fixing the pre-trained model gets even worse performance on general test set \cite{sathyendra2022contextual}. We propose to use a group-based learning rate policy, and achieved better performance on both biased and unbiased words.

\subsection{Text-only ASR}
Due to the sparsity of audio training data, additional text data is explored to improve the accuracy of rare words. Several studies focus on leveraging external knowledge to enrich the representations of rare words, such as selecting text data for LM training \cite{huang2022sentence}, augmenting the rare word embedding to enhance LM's performance \cite{khassanov2019enriching,huang2021lookup}. 
Nevertheless, all these methods require an external LM during inference, which incurs computational cost.

Other methods try to increase the accuracy of rare words through joint training with text-only data \cite{sainath2023improving,chen2022maestro,meng2023jeit}. However, these works focus on enhancing the general performance of rare words, instead of particular biasing words.
In contrast, we propose to employ USTR, which is adopted for text-only domain adaptation \cite{huang2023text}, to further enhance the capability of deep biasing with more unpaired text data.

\section{Proposed Methods}
\label{sec:method}

\subsection{Model architecture}
\label{model_architecture}

The overall architecture is illustrated in Figure \ref{fig:model}(a), which consists of \texttt{Transducer}, USTR's \texttt{TextEncoder} and \texttt{BiasingModule}. The \texttt{BiasingModule} includes two parts, named \texttt{WordEncoder} and \texttt{BiasingLayer}. 

For the paired speech and text data, let $\textbf{X}^{\text{speech}}\in {\mathbb{R}}^{B_1\times T \times D_1}$ be the audio features like Fbank, and the output of encoder is computed by
\begin{equation}
    \textbf{H}^{\text{speech}}=\texttt{Encoder}(\textbf{X}^{\text{speech}}),
\end{equation}
where $\textbf{H}^{\text{speech}} \in {\mathbb{R}}^{B_1\times T' \times H}$, and $\texttt{Encoder}(\cdot)$ is the same as $\texttt{SharedEncoder}(\texttt{AudioEncoder}(\cdot))$.

For unspoken text data, let $\textbf{X}^{\text{text}}\in \mathbb{R}^{B_2\times N \times D_2}$ be the text features, and the output of encoder is computed by
\begin{equation}
    \textbf{H}^{\text{text}}=\texttt{SharedEncoder}(\texttt{TextEncoder}(\textbf{X}^{\text{text}})),
\end{equation}
where $\textbf{H}^{\text{text}} \in {\mathbb{R}}^{B_2\times N' \times H}$.

Then the encoder output of paired speech-text data $\textbf{H}^{\text{speech}}$ and unspoken text data $\textbf{H}^{\text{text}}$ are concatenated on the batch dimension by filling to the same size on length dimension, as well as the  output label sequence,
\begin{equation}
\textbf{H}^{\text{enc}}=\texttt{BatchConcat}(\textbf{H}^{\text{speech}}, \textbf{H}^{\text{text}}),
\end{equation}
\begin{equation}
\textbf{Y}=\texttt{BatchConcat}(\textbf{Y}^{\text{speech}}, \textbf{Y}^{\text{text}}),
\end{equation}
where $\textbf{H}^{\text{enc}}\in {\mathbb{R}}^{B \times L \times H}$, $B=B1+B2$, $L=\text{max}(T',N')$, $\textbf{Y}\in {\mathbb{R}}^{B\times U}$.

For simplicity, let $\textbf{y}\in {\mathbb{R}}^{U}$ and $\textbf{h}^{\text{enc}} \in \mathbb{R} ^ {L \times H}$ be an utterance in the batch of $\textbf{H}^{\text{enc}}$ and $\textbf{Y}$, and the predicted probability on vocabulary of CT at frame $t$ and step $u$ is computed by
\begin{equation}
    \textbf{h}^{\text{pred}}_{u}=\texttt{Predictor}(\textbf{y}_{0:u-1}),
\end{equation}
\begin{equation}
    \textbf{h}^{\text{joint}}_{t,u}=\texttt{Jointer}(\textbf{h}^{\text{enc}}_{t},\textbf{h}^{\text{pred}}_{u}),
\end{equation}
\begin{equation}
    \hat{\textbf{y}}_{t,u}=\texttt{Softmax}(\texttt{FC}(\textbf{h}^{\text{joint}}_{t,u})).
\end{equation}
Then with forward-backward algorithm \cite{graves2012sequence}, Transducer loss is computed as the training objective function.

For \texttt{BiasingModule}, it takes the bias words $\textbf{W}_{1:K}\in \mathbb{R} ^ {K\times S}$ as input and converts $\textbf{W}_{1:K}$ to $\textbf{E}_{1:K}\in \mathbb{R} ^ {K\times M}$ by a \texttt{WordEncoder}, where $K$ is the number of biasing words (including one empty word for no biasing), $S$ is the max length of rare words, $M$ is the dimension of word embedding, $\textbf{W}_k$ is a word or word sequence, and $\textbf{E}_k$ is the corresponding embedding. More details are provided in Section \ref{subsec:phone}.

\texttt{BiasingLayer} contains multi-head attention (MHA), which takes the input from Transducer/USTR as query and $\textbf{E}_{1:K}$ as key/value. The output of MHA is reshaped to the same size as the query by a projection layer, then is added to the original query as a biasing vector. More details about \texttt{BiasingLayer} can be found in Section \ref{subsec:query}.

\begin{figure}[t]
    \centering
    \includegraphics[width=0.49\textwidth]{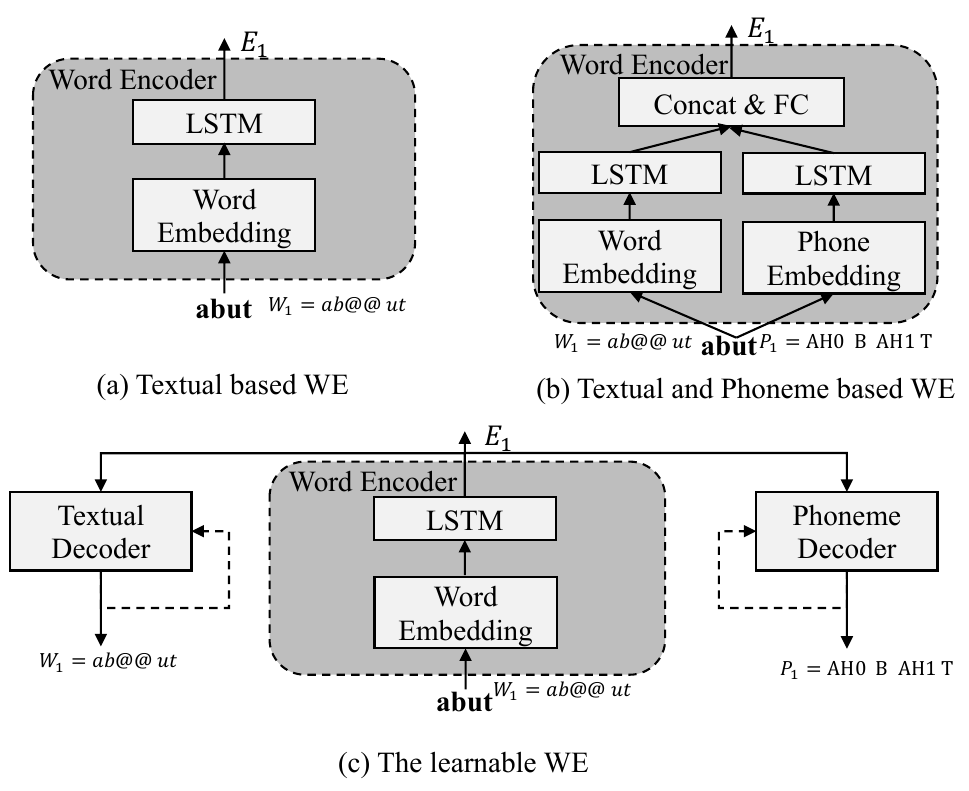}
    \caption{The model structures of different word encoders (WE). Only one word \textbf{\texttt{abut}} is considered here for better understanding, where "ab@@ ut" is the sub-word sequence and "AH0 B AH1 T" is the phoneme sequence.}
    \label{fig:word_encoder}
\end{figure}

\subsection{Biasing with different queries}
\label{subsec:query}

\textbf{Predictor-Query} (Figure \ref{fig:model}(b)). The query is the concatenated value of predictor's embedding output $\textbf{h}^{\text{emb}_u}$ and final output $\textbf{h}^{\text{pred}_u}$ at each step $u$, and the biasing process is
\begin{equation}
\text{Query}=\texttt{Concat}(\textbf{h}^{\text{emb}}_u, \textbf{h}^{\text{pred}}_u),
\end{equation}
\begin{equation}
\textbf{b}_u=\text{MHA}(\text{Query}, \textbf{E}_{1:K}, \textbf{E}_{1:K}),
\end{equation}
\begin{equation}
\hat{\textbf{h}}^{\text{pred}}_u = \textbf{b}_u + \textbf{h}^{\text{pred}}_u,
\end{equation}
where $\hat{\textbf{h}}^{\text{pred}}_u$ will replace $\textbf{h}^{\text{pred}}_u$ for Transducer/USTR training and inference.

\textbf{Encoder-Query} (Figure \ref{fig:model}(c)). The query is the encoder output $\textbf{h}^{\text{enc}}_{t}$ at each time step $t$, and the biasing  process is
\begin{equation}
\textbf{b}_t=\text{MHA}(\text{Query}=\textbf{h}^{\text{enc}}_{t}, \textbf{E}_{1:K}, \textbf{E}_{1:K})
\end{equation}

\textbf{Enc-Pre Query}. This is the combination of Encoder-Query and Predictor-Query. As noted in Figure \ref{fig:model}(d), there are two \texttt{BiasingLayer} modules with separated parameters.

\textbf{Jointer-Query} (Figure \ref{fig:model}(e)). In this case, the query is the hidden states $\textbf{h}^{\text{joint}}_{t,u}$ in jointer, and the biasing  process is
\begin{equation}
\textbf{b}_{t,u}=\text{MHA}(\text{Query}=\textbf{h}^{\text{joint}}_{t,u}, \textbf{E}_{1:K}, \textbf{E}_{1:K})
\end{equation}

\subsection{Combining textual and phoneme features}
\label{subsec:phone}

\textbf{Textual-WE}, as shown in \ref{fig:word_encoder}(a). A word is first converted to a sub-word sequence and an embedding layer is used to extract token embedding. Then the sequence of token embedding is fed into an unidirectional long short-term memory (LSTM) layer, and the final state is adopted as the word embedding.

\textbf{Tex-Pho-WE}, as illustrated in \ref{fig:word_encoder}(b). Different from Textual-WE, there is another branch, which converts the word to a phoneme sequence, and the sequence of phoneme embedding is fed to the LSTM layer to get the final state as a  phoneme representation of the word. The textual and phoneme representations are concatenated and reshaped to the same size by a fully connected (FC) layer.

\textbf{Learnable-WE}, as illustrated in \ref{fig:word_encoder}(c). Different from Textual-WE, there are two additional transformer decoders, which take the word embedding as input and predict textual and phoneme sequences respectively. The Learnable-WE is similar to that in \cite{chen2019joint}. It should be noticed that the decoders can be removed during inference.

\section{Experimental Setup}
\label{sec:exp_set}
\subsection{Data sets}

The experiments are conducted on LibriSpeech \cite{panayotov2015librispeech} corpus, where the 960-hour audio data
is adopted as paired speech-text corpus, and the normalized text data with size of 1.5G\footnote{\url{https://www.openslr.org/11/}} is used as unspoken text for training USTR. Also, in this work, we proposed to use the 209.2k rare words (defined as words not in the 5,000 most common words in the paired audio training set, i.e., \textbf{Rare5k}) as unspoken text to improve the biasing performance when training with USTR. 

For the paired audio data, 3-fold speed perturbation \cite{ko2015audio} with factors of 0.95, 1.0, and 1.05 is used for data augmentation. Besides, the 80-dim filter-bank (Fbank)  is extracted and Spec-Augment \cite{park2019specaugment} is applied on Fbank features before feeding into \texttt{AudioEncoder}.

When training USTR with unspoken text data, the phonemes are adopted as text features, which is similar to that in \cite{huang2023text}, and the text features are masked with a probability of 0.15 before repeating and feeding into \texttt{TextEncoder}.

\subsection{Model}
The model's structure is described as that in Section \ref{model_architecture}, where \texttt{AudioEncoder} consists of 2-layer 2D convolution with channel=128, kernel=3, stride=2 and ReLU activation \cite{glorot2011deep, dahl2013improving}, resulting in downsampling of 40ms. \texttt{TextEncoder} contains an embedding layer and a Transformer layer. And \texttt{SharedEncoder} consists of 12 streaming Conformer \cite{gulati2020conformer} layers, where the attentions of first 7 layers have a look ahead of 1 frame (i.e., 40ms) and there is no look ahead for all convolutions and attentions of last 5 layers. The total look ahead of the \texttt{Encoder} is 310ms (280ms for Conformer layers and 30ms for \texttt{AudioEncoder}). \texttt{Predictor} has an embedding layer and 2 LSTM layers and \texttt{Jointer} has a linear layer. The output of RNN-T is 4,048 subword units \cite{sennrich2016neural}.
All models are implemented and trained with PyTorch \cite{paszke2019pytorch}.

\begin{table}[t]
  \caption{The WER(U-WER/B-WER)(\%) results of deep biasing with different queries. The size of bias list is 100 here.}
  \label{tab:baseline}
  \centering
  \begin{tabular}{|l|c|c|}
    \hline
    model  & test-clean &  test-other \\
    \hline
    \hline
    \multirow{2}{*}{CT Baseline} &3.28 & 7.88 \\
    & (2.15/12.43) & (5.71/26.97) \\
    \hline
    \ + deep biasing & 2.93 & 7.11 \\
    \ (Predictor-Query) & (2.16/9.22) & (5.61/20.36)  \\
    \hline
    \ + deep biasing & 2.78 & 6.63 \\
    \ (Encoder-Query) &  (2.11/8.18) & (5.38/17.66) \\
    \hline
    \ + deep biasing & \textbf{2.67} & \textbf{6.54} \\
    \ (Enc-Pre Query) & (\textbf{2.06}/7.64) & (\textbf{5.48/15.81}) \\
    \hline
    \ \multirow{2}{*}{\ \ + Freezing CT} & 2.92 & 7.01 \\
    \ & (2.14/9.27) & (5.61/19.31) \\
    \hline
    \ + deep biasing & \textbf{2.67} & 6.67 \\
    \ (Jointer-Query) & (2.07/\textbf{7.50}) & (5.51/16.88) \\
    \hline
  \end{tabular}
\end{table}

\subsection{Training}

During training, the bias list of current batch is extracted from all the batch references, including the rare words in \textbf{Rare2k} (defined as words that fall outside the 2k most common words in the paired audio training set).

The textual feature of a bias word is the subword units. Besides, the phoneme features are generated by a Grapheme-to-Phoneme system, i.e., \texttt{g2pE}\footnote{\url{https://github.com/Kyubyong/g2p}}.

When training USTR with unspoken text data, single-step is adopted, as the training process is simpler and more efficient, and better performance can be obtained \cite{huang2023text}. Besides, the USTR is trained from scratch with \texttt{BiasingModule}. Also, during the training of USTR, paired speech-text data is fed into the \texttt{TextEncoder} by using text features instead of audio features with a probability 0.15 to force the \texttt{TextEncoder} and \texttt{AudioEncoder} to learning an unified representation for audio and text features.

However, when training \texttt{BiasingModule} with pared speech-text data, a pre-trained Conformer Transducer (CT) is used for initialization. In this case, to maintain the model's performance when there is no biasing, group-based learning rate (lr) policy is proposed, where \texttt{BiasingModule} and CT are trained with lr=1e-5 and lr=1e-7 jointly.

In addition to the Transducer loss, CTC and an extra AED decoder are adopted for multi-task learning. Also, internal LM estimation (ILMT) loss is chosen as an auxiliary loss. The overall training loss is
\begin{equation}
    \mathcal{L}=\mathcal{L}_{\texttt{Transducer}} + \mathcal{L}_{\texttt{CTC}} + \mathcal{L}_{\texttt{AED}} + \lambda\mathcal{L}_{\texttt{ILMT}},
\end{equation}
where $\lambda$ is set to 0.2 in all experiments. When Learnable-WE is used, there still exists two AED losses with scale of 0.1. Besides, subword regularization, i.e., BPE dropout \cite{provilkov2020bpe}, is applied with a probability of 0.1 during training.

\subsection{Inference}
During inference, for each utterance, the biasing list is constructed by extracting the rare words (in \textbf{Rare5k}) in the reference and adding a certain number of distractors.
The biased words with different size (100/500/1000/2000)
are explored to check the performance of proposed method with different scale of bias list size, which is the same as that in \cite{le2021contextualized}.

WER is evaluated on Librispeech \texttt{test-clean} and \texttt{test-other} sets. To indicate the performance of (biased) rare words, B-WER (biased WER) is measured on words in the biasing list as that in \cite{le2021contextualized,huang2023contextualized}, while U-WER (unbiased WER) is also measured to prevent degrading the performance of words not in the biasing list.

Moreover, similar to that in \cite{sun2021tree}, rare WER (R-WER) is used to evaluate the performance of proposed methods for utterance-level, chapter-level and book-level biasing.

\section{Experimental Results}
\label{sec:exp_res}
\subsection{Biasing with different queries}

Deep biasing with different queries in CT has been initially explored, and the results are illustrated in Table \ref{tab:baseline}. Compared to the CT baseline, all models with deep biasing achieve significant reductions on B-WER, and the U-WER performances remain almost the same or even better. It indicates that the proposed method enhances the performance of rare words without sacrificing the performance of common words.


Joint Query achieves the best B-WER on test-clean, while Enc-Pre Query obtains better WERs on test-other. When applying deep biasing with Enc-Pre Query, we observe relative improvements of 38.5\% and 41.4\% on the B-WER. However, when we tried to freeze the parameters of CT rather than using different learning rates, the performance becomes much worse on both biased and unbiased words.

Besides, for an utterance in which lengths of encoder's output and predictor's output are $L$ and $U$, the computational complexity of Enc-Pre Query is $O(L+U)$, in contrast the computational complexity of Jointer Query is $O(L\times U)$, which is larger when $U\geq2$, $L > U$. Therefore, Enc-Pre Query has a lower computational complexity in most cases and is chosen as the default in the following experiments.

\subsection{Combining textual and phoneme features}

We evaluate the effectiveness of combining textual and phoneme information with different word encoders. As shown in Figure \ref{fig:word_encoder}, compared with Textual-WE, Tex-Pho-WE obtained the best results, with relative reductions of 10.47\% (7.64\% $\rightarrow$ 6.84\%) and 7.08\% (15.81\% $\rightarrow$ 14.69\%) on the B-WERs of two test sets, respectively. Learnable-WE, explored in \cite{chen2019joint}, performs worse than Tex-Pho-WE, as Tex-Pho-WE uses the phoneme information without errors. The improvement indicates the benefits of phoneme information for deep biasing in CT, and Tex-Pho-WE is chosen as the default configuration in following experiments.

\begin{table}[t]
  \caption{The WER(U-WER/B-WER)(\%) results of deep biasing when using difference embedding modules. The size of bias list is 100, and Enc-Pre Query is used for all experiments.}
  \label{tab:phone}
  \centering
  \begin{tabular}{|l|c|c|}
    \hline
    model & test-clean & test-other \\
    \hline
    \hline
    \multirow{2}{*}{CT Baseline} & 3.28 & 7.88 \\
    & (2.15/12.43) & (5.71/26.97) \\
    \hline
    deep biasing & 2.67 & 6.54 \\
    \ \ \ (Textual-WE) & (2.06/7.64) & (5.48/15.81) \\
    \hline
    deep biasing & \textbf{2.56} & \textbf{6.33} \\
    \ \ \ (Tex-Pho-WE) & (\textbf{2.03/6.84}) & (\textbf{5.38/14.69}) \\
    \hline
    deep biasing & 2.68 & 6.44 \\
    \ \ \ (Learnable-WE) & (2.09/7.50) & (5.40/15.55) \\
    \hline
  \end{tabular}
\end{table}

\begin{table}[!ht] 
  \caption{The WER(U-WER/B-WER)(\%) results when combining deep biasing with USTR. The size of bias list
is 100 here. USTR-CT(C/L) denotes the USTR model trained using Librispeech LM \textbf{C}orpus and rare word \textbf{L}ist respectively.}
  \label{tab:ustr}
  \centering
  \begin{tabular}{|l|c|c|}
    \hline
    model & test-clean &  test-other \\
    \hline
    \hline
    \multirow{2}{*}{CT Baseline} & 3.28 & 7.88 \\
    & (2.15/12.43) & (5.71/26.97) \\
    \hline
    \multirow{2}{*}{\ \ + deep biasing} &   2.56  & 6.33    \\
    & (2.03/6.84) & (5.38/14.69) \\
    \hline
    \hline
    \multirow{2}{*}{USTR-CT(C)} &  3.05   &  7.49  \\
    & (2.09/10.83) & (5.55/24.58) \\
    \hline
    \multirow{2}{*}{\ \ + deep biasing} &  2.39  & 6.30    \\
    & (2.27/3.38) & (6.23/6.99) \\
    \hline
    \hline
    \multirow{2}{*}{USTR-CT(L)} &  3.13  & 7.58  \\
    & (2.06/11.84) & (5.57/25.31) \\
    \hline
    \multirow{2}{*}{\ \ + deep biasing} &  2.19  & 5.61    \\
    & (\textbf{1.99}/3.82) & (\textbf{5.38}/7.57) \\
    \hline
    \hline
    \multirow{2}{*}{USTR-CT(C+L)} &  2.98  & 7.45 \\
    & (1.97/11.14) & (5.54/24.24) \\
    \hline
    \multirow{2}{*}{\ \ + deep biasing} &  \textbf{2.15}   &  \textbf{5.56}   \\
    & (2.00/\textbf{3.33}) & (5.46/\textbf{6.45}) \\
    \hline
  \end{tabular}
\end{table}

\subsection{Combined with USTR}
\label{subsection_ustr}

\begin{table*}[th]
    \caption{The WER(U-WER/B-WER)(\%) results on LibriSpeech test sets with different bias list size (100/500/1000/2000).}
    \label{tab:list_size}
    \centering
    \begin{threeparttable} 
    \begin{tabular}{|l|cc|cc|cc|cc|}
        \hline
        \multirow{2}{*}{Method} & \multicolumn{2}{c|}{$N=100$} & \multicolumn{2}{c|}{$N=500$} & \multicolumn{2}{c|}{$N=1000$} & \multicolumn{2}{c|}{$N=2000$} \\  
         & test-clean & test-other & test-clean & test-other & test-clean & test-other & test-clean & test-other \\
        \hline
        \hline
        \multirow{2}{*}{CT Baseline} & 3.28 & 7.88 & 3.28 & 7.88 & 3.28 & 7.88  & 3.28 & 7.88 \\
        & (2.2/12.4) & (5.7/27.0) &(2.2/12.4) & (5.7/27.0) & (2.2/12.4) & (5.7/27.0) & (2.2/12.4) & (5.7/27.0)  \\
        \hline
        \multirow{2}{*}{Enc-Pre Query}  & 2.67 & 6.54 & 2.87 &  7.01 & 2.97 & 7.30 & 3.09 & 7.44 \\
        \ \ & (2.1/7.6) & (5.5/15.8) & (2.1/9.1) & (5.5/20.3) & (2.1/10.0) & (5.6/22.3) & (2.2/10.8) & (5.6/23.5) \\
        \hline
        \multirow{2}{*}{+ Tex-Pho-WE} & 2.56 & 6.33 & 2.74 &  6.70 & 2.81 & 6.93 & 2.91 & 7.09 \\
         & (2.0/6.8) & (5.4/14.7) & (2.1/8.1) & (5.5/17.5) & (2.1/8.7) & (5.5/19.1) & (2.1/9.5) & (5.6/20.6) \\
        \hline
        \multirow{2}{*}{\ \ + USTR(C+L)} & 2.15 & 5.56 & 2.23 & 5.83  & 2.28 & 6.01 & 2.30 & 6.14 \\
        & (2.0/3.3) & (5.5/6.5) & (2.1/3.7) & (5.6/8.2) & (2.1/3.8) & (5.7/9.1) & (2.1/4.4) & (5.6/11.0) \\
        \hline
        \multirow{2}{*}{\ \ \ \ + FST} & 2.06 & \textbf{5.38} & \textbf{2.09} & \textbf{5.62} & 2.16 & \textbf{5.75} & \textbf{2.17} & \textbf{5.84} \\
        &(2.1/\textbf{2.0}) & (5.5/\textbf{4.4}) & (2.1/\textbf{2.2}) & (5.6/\textbf{5.6}) & (2.1/\textbf{2.5}) & (5.7/\textbf{6.3}) & (2.1/\textbf{3.0}) & (5.6/\textbf{7.6}) \\
        \hline
        \hline
        DB-RNN-T + FST & \textbf{1.98} & 5.86 & \textbf{2.09} & 6.09  & \textbf{2.14} & 6.35 & 2.27 & 6.58 \\
        \ \ \ \ + DB-NNLM\cite{le2021contextualized}& (\textbf{1.5}/5.7) & (\textbf{4.9}/14.1) & (\textbf{1.6}/6.2) & (\textbf{5.1}/15.1) & (\textbf{1.6}/6.7) & (\textbf{5.1}/17.2) & (\textbf{1.6}/7.3) & (\textbf{5.2}/18.9) \\
        \hline
        \hline
        CT + deep & 3.66 & 7.63 & 3.78 & 7.99 & 3.88 & 8.28 &  \multirow{2}{*}{N/A} & \multirow{2}{*}{N/A}  \\
        \ \ \ \ biasing\cite{huang2023contextualized}& (2.8/11.2) & (6.0/22.1) & (2.9/11.5) & (6.2/23.4) & (2.9/11.9) & (6.4/24.5) & & \\
        \hline
    \end{tabular}
    \begin{tablenotes}
    \footnotesize
        \item[*] \textbf{N/A} means that the results are not available.
    \end{tablenotes}
    \end{threeparttable}
\end{table*}

\begin{table*}[!ht] 
    \caption{The WER/R-WER(\%) results of various systems on LibriSpeech test sets when using deep bias with utterance-level, chapter-level and book-level rare words. The bias list size is 1000 for all methods.}
    \label{tab:comp_tcp_gen}
    \centering
    \begin{threeparttable} 
    \begin{tabular}{|l|ccc|ccc|}
        \hline 
        \multirow{2}{*}{Model} & \multicolumn{3}{c|}{test-clean} & \multicolumn{3}{c|}{test-other } \\
        \cline{2-7} & Utterance-level & Chapter-level & Book-level & Utterance-level & Chapter-level & Book-level \\
        \hline
        \hline
        RNN-T + TCPGen\cite{sun2021tree} & 4.9(13.9) & 5.1(13.6) & 5.4(28.2) & 14.0(35.0) & 14.1(32.4) & 14.8(52.1) \\
        \ \ + deep biasing + SF & 3.8(11.3) & 4.0(11.0) & 4.2(24.0) & 11.5(29.0) & 12.0(29.3) & 12.2(50.8) \\
        \hline
        \hline
        DB-RNN-T\cite{le2021contextualized}\tnote{*} & 3.3(11.9)& N/A& N/A & 9.1(31.4) & N/A & N/A \\
        \ \ + FST + DB-NNLM\tnote{*} & \textbf{2.1}(6.8)& N/A & N/A & 6.4(21.3) & N/A & N/A \\
        \hline
        \hline
        Proposed deep biasing & 2.8(9.8) & 3.0(10.9) & 3.2(12.9) & 6.9(22.9) & 7.2(25.4) & 7.6(29.3)\\
        \ \ + USTR + FST & 2.2(\textbf{5.0})& \textbf{2.5}(\textbf{7.1}) & \textbf{2.8}(\textbf{9.8})& \textbf{5.8}(\textbf{11.7}) & \textbf{6.4}(\textbf{18.0}) & \textbf{7.1}(\textbf{24.2}) \\
        \hline
    \end{tabular}
    \begin{tablenotes}
    \footnotesize
        \item[*] The results of R-WER is generated by using the hypothesis files in \url{https://github.com/facebookresearch/fbai-speech/tree/main/is21_deep_bias} with a bias list size of 1000.
    \end{tablenotes}
    \end{threeparttable} 
\end{table*}

Then, we investigate the impact of introducing unpaired text data containing rare words by combining deep biasing and USTR. As illustrated in Table \ref{tab:ustr}, compared to CT baseline, not only the USTR with LM corpus obtains better performance on all WERs, but also the USTR with rare word list obtains slight improvements. When trained using both LM corpus and rare word list, the B-WERs on test-clean and test-other are reduced from 12.43\%/26.97\% to 11.14\%/24.24\%.

When combining USTR with deep biasing, significant improvements are observed. When trained with unpaired text data C/L, the B-WER of deep biasing model on test-other is improved from 14.69\% to 6.99\%/7.57\%, much better than CT or USTR baselines. 
The improvements are mainly attributed to utilization of more rare words for training and the capacity of biasing module capacity is enhanced.

It should be noted that the rare words list contains only 209.2k rare words, which is significantly smaller than the LM corpus and demonstrates the feasibility of our method. Consequently, we combined the LM corpus and rare word list, and achieved the best WER/B-WER on both test sets. Compared with the deep biasing baseline, USTR(C+L) with deep biasing provides relative improvements of 51.32\%/56.09\% on B-WER (6.84\%/14.69\% $\rightarrow$ 3.33\%/6.45\%).

\subsection{Different bias list size}
We further evaluate the robustness of the proposed method on large-scale bias lists, in which most words are irrelevant to the audio. As illustrated in Table \ref{tab:list_size}, as the size of bias list increases, B-WER increases gradually when using Enc-Pre Query. 
The absolute gaps of B-WER on the two test sets between $N=100/2000$ are 3.2\%/7.7\% respectively.
Tex-Pho-WE alleviates these gaps to 2.7\%/5.9\% because it provides additional information to discriminate similar grapheme sequences.
By combining USTR(C+L), the gap shrinks to 1.1\%/4.5\%. Compared to CT baseline, our best system, which combines deep biasing, USTR, and FST, achieves relative B-WER reductions of 75.81\% and 71.85\% on the two test sets respectively when $N=2000$.

\subsection{Comparison with other methods}
Results of some prior works are also listed in the bottom rows in Table \ref{tab:list_size}. Compared to the best system in \cite{le2021contextualized}, the proposed method achieves the best B-WER on all test sets and all sizes of bias list, while no external LM is used. With an external LM, we believe further improvements can be achieved. 

R-WERs of our system and other approaches are listed in Table \ref{tab:comp_tcp_gen} with utterance-level, chapter-level, and book-level rare words as those in \cite{sun2021tree}. The proposed strategy achieves the best R-WERs on two test sets with all levels of the bias list, indicating the proposed method's superiority.

\section{Conclusions}
\label{sec:con}
In this paper, we proposed several significant improvements in large-scale deep biasing for Transducer based streaming ASR. Our approach extends CT by incorporating textual and phoneme information of rare words, resulting in notable relative improvements of 45.16\% and 45.56\% on B-WER over the baseline.
Furthermore, by incorporating the previously established USTR method for text injection during training and incorporating FST during inference, the proposed approach yields remarkable improvements, leading to relative reductions of 83\% $\sim$ 84\% on B-WER. Moreover, our method demonstrates robustness in large-scale deep biasing scenarios, effectively closing the gap between bias list sizes from 100 to 2000. Notably, compared to other publicly available results, our approach attains state-of-the-art performance on the accuracy of rare words.



\bibliographystyle{IEEEbib}
\bibliography{strings,refs}

\end{document}